\documentclass{article}

\usepackage{microtype}
\usepackage{graphicx}
\usepackage{subcaption}
\usepackage{booktabs} 
\usepackage{multirow}
\usepackage{hyperref}

\usepackage[table]{xcolor}

\usepackage[preprint]{icml2026}

\usepackage{amsmath}
\usepackage{amssymb}
\usepackage{mathtools}
\usepackage{amsthm}

\usepackage[capitalize,noabbrev]{cleveref}

\theoremstyle{plain}
\newtheorem{theorem}{Theorem}[section]

\newtheorem{lemma}[theorem]{Lemma}

\theoremstyle{definition}

\theoremstyle{remark}

\usepackage[textsize=tiny]{todonotes}

\icmltitlerunning{Parameter-Efficient Fine-Tuning of LLMs with Mixture of Space Experts}

\begin{document}

\pagestyle{empty}
\thispagestyle{empty}

\twocolumn[
  \icmltitle{Parameter-Efficient Fine-Tuning of LLMs with Mixture of Space Experts}



  \icmlsetsymbol{equal}{*}

  \begin{icmlauthorlist}
    \icmlauthor{Buze Zhang}{xjtu}
    \icmlauthor{Jinkai Tao}{cufe}
    \icmlauthor{Zilang Zeng}{hkust}
    \icmlauthor{Neil He}{uiuc}
    \icmlauthor{Ali Maatouk}{yale}
    \icmlauthor{Menglin Yang}{hkust}
    \icmlauthor{Rex Ying}{yale}
  \end{icmlauthorlist}
  
\icmlaffiliation{yale}{Department of Computer Science, Yale university, New Haven, CT, USA}
\icmlaffiliation{uiuc}{Department of Computer Science, University of Illinois Urbana-Champaign, Urbana, IL, USA}
\icmlaffiliation{hkust}{Department of Artificial Intelligence, Hong Kong University of Science and Technology (Guangzhou), Guangzhou, Guangdong, China}
\icmlaffiliation{xjtu}{School of Software Engineering, Xi’an Jiaotong University, China}
\icmlaffiliation{cufe}{School of Information, Central University of Finance and Economics, Beijing, China}

\icmlcorrespondingauthor{Menglin Yang}{menglin.yang@outlook.com}
\icmlkeywords{Machine Learning, ICML}





  \vskip 0.3in
]



\printAffiliationsAndNotice{}  

\begin{abstract}
Large Language Models (LLMs) have achieved remarkable progress, with Parameter-Efficient Fine-Tuning (PEFT) emerging as a key technique for downstream task adaptation. However, existing PEFT methods mainly operate in Euclidean space, fundamentally limiting their capacity to capture complex geometric structures inherent in language data. 
While alternative geometric spaces, like hyperbolic geometries for hierarchical data and spherical manifolds for circular patterns, offer theoretical advantages, forcing representations into a single manifold type ultimately limits expressiveness, even when curvature parameters are learnable. To address this, we propose \textbf{Mixture of Space} (\textbf{MoS}), a unified framework that leverages multiple geometric spaces simultaneously to learn richer, curvature-aware representations. 
Building on this scheme, we develop \textbf{MoSLoRA}, which extends Low-Rank Adaptation (LoRA) with heterogeneous geometric experts, enabling models to dynamically select or combine appropriate geometric spaces based on input context. 
Furthermore, to address the computational overhead of frequent manifold switching, we develop a lightweight routing mechanism. Moreover, we provide empirical insights into how curvature optimization impacts training stability and model performance.
Our experiments across diverse benchmarks demonstrate that MoSLoRA consistently outperforms strong baselines, achieving up to 5.6\% improvement on MATH500 and 15.9\% on MAWPS.
\end{abstract}

\section{Introduction}

Large Language Models (LLMs) have recently demonstrated impressive performance across a wide range of applications, including translation, comprehension, dialogue, and reasoning \citep{achiam2023gpt,jaech2024openai,dubey2024llama,team2024qwen2}. With the aid of post-training techniques such as instruction tuning, they can be further adapted to diverse downstream tasks with notable gains in effectiveness \citep{hu2023llm,han2024parameter}. 
Despite these advances, most existing approaches rely on a Euclidean assumption, modeling all embeddings in flat Euclidean space, which are often inadequate to capture the semantic diversity and contextual complexity of natural language \citep{bronstein2017geometric,park2024geometry,he2025position}.

Semantic structures in language often display geometric patterns: hierarchical relationships where broad and general concepts naturally encompass finer subcategories and distinct entities; circular patterns among synonymous expressions or co-referential terms (such as the days of the week, e.g., Monday, ..., Sunday, and months of a year) \citep{engels2024not}; and complex multi-level dependencies that resist simple linear organization. These patterns are largely overlooked and constrained under Euclidean representations, leaving open the question of \textit{how to effectively leverage such naturally occurring structures within embedding spaces to unlock richer representational capacity?}

Recently, growing attention has shifted toward non-Euclidean constant-curvature spaces as alternatives to Euclidean space for improving model performance \citep{peng2021hyperbolic,yang2024hypformer,pal2024compositional,loshchilov2024ngpt}. From an embedding perspective, it has been observed that tokens associated with higher-level and more general semantics often occupy regions of lower norm, whereas tokens tied to more concrete and specific meanings are distributed in regions of higher norm \citep{yang2024hyperbolic}. Hyperbolic space, with its negative curvature and exponential growth capacity, offers an effective means of embedding complex hierarchical information in lower dimensions compared to Euclidean space. Building on these advantages, \citet{yang2024hyperbolic} explored combining LoRA with hyperbolic geometry, enabling efficient fine-tuning of pretrained LLMs while reducing embedding distortion. Similarly, spherical manifolds have shown promise for capturing intrinsic circular patterns and normalized representations, as demonstrated by \citet{loshchilov2024ngpt} who reformulated Transformers as hyperspherical models (nGPT) by enforcing embeddings to lie on the unit sphere.

\textbf{Limitations of Existing Methods.} However, real-world language data exhibits complex, heterogeneous structural relationships that cannot be adequately captured by constraining representations to a single geometric space. For instance, a single sentence may contain both hierarchical semantic relationships (e.g., category-subcategory structures) and circular patterns (e.g., synonymous expressions or co-referential terms), requiring different geometric inductive biases simultaneously. Furthermore, existing non-Euclidean approaches face significant computational challenges. Prior work often employs exponential and logarithmic maps to transition between non-Euclidean and Euclidean spaces \citep{ganea2018hyperbolic,yang2022hyperbolic}, incurring substantial computational overhead at each model layer. These repeated mappings are difficult to scale to larger models and greater depth, creating a practical barrier to widespread adoption.

To address these fundamental limitations, we propose a unified \textbf{Mixture of Space (MoS)} scheme that integrates three types of constant-curvature spaces (hyperbolic, spherical, and euclidean), capturing diverse geometric structures within a single model simultaneously. Rather than constraining all representations to a single geometric paradigm, our approach allows different tokens to reside in the geometric space most suited to their structural properties: hierarchical concepts in hyperbolic space, circular patterns in spherical space, and flat relational structures in Euclidean space. 

Building upon this framework, we introduce \textbf{MoSLoRA}, which combines the MoS paradigm with Low-Rank Adaptation (LoRA) for efficient fine-tuning of large language models. MoSLoRA employs a lightweight token routing mechanism that dynamically assigns each token to its optimal geometric expert, avoiding the computational overhead of repeated space transformations while maintaining the representational benefits of multiple geometries. This design enables the model to adapt its geometric inductive biases on-the-fly, matching the heterogeneous structural requirements of real-world language data. Our contributions can be summarized as follows: 
\begin{itemize}
\item We introduce a unified architecture that integrates three distinct constant-curvature spaces, and combine it with Mixture-of-Experts (MoE) and LoRA to form a novel and efficient fine-tuning framework for LLMs.
\item We design a lightweight token routing mechanism that efficiently directs tokens among geometric spaces to overcome high-overhead space transformation.
\item We provide an in-depth analysis of the training dynamics of space selection and routing strategies, along with optimizing geometric space integration during fine-tuning.
\item We evaluate the proposed method on benchmarks including natural language understanding and mathematical reasoning, where it consistently outperforms several strong baselines.
\end{itemize}

\section{Related Work}

\subsection{Mixture of LoRA Experts}
MoE introduces multiple expert networks and a gating network that selects experts based on different data characteristics \citep{Jacobs_Jordan_Nowlan_Hinton_1991, shazeer2017}, to improve the capacity and computational efficiency of large-scale models through combinations of experts. 

Recent work has begun to equip PEFT with MoE-style sparsity, yielding a line of LoRA-based mixture methods \citep{peft, gao2022parameter, zadouri2024pushing, feng2024mixture, wu2024mixture, wu-etal-2024-mixture-subspaces}. Representative ones include MELoRA \citep{ren2024melora}, which constructs multiple LoRA experts by diagonal partitioning, HydraLoRA \citep{tian2024hydralora}, which shares one LoRA factor ($W_A$) across experts while keeping multiple expert-specific factors ($W_B$), and HMoRA \citep{liao2025hmora}, which introduces hierarchical routing for multi-task settings. 
Despite their gains, these methods typically rely on homogeneous LoRA experts and often face a practical trade-off between performance and efficiency: stronger accuracy tends to require activating more experts (or higher effective rank), which increases compute and dilutes the sparsity benefits. 
In contrast, we extend LoRA experts to heterogeneous geometric spaces, improving expressiveness per activated parameter and thus achieving a better accuracy - efficiency balance.

\subsection{Non-Euclidean and Curvature-Aware Modeling}
Recent research increasingly emphasized the importance of non-Euclidean geometry for representation learning, aiming to overcome the limitations of standard Euclidean embedding spaces \citep{shimizu2020hyperbolic,peng2021hyperbolic,pal2024compositional,he2025lresnet,he2025position}. Early attempts focused on constructing neural networks that operated fully in hyperbolic space. For example, \citet{chen2022fully} formalized all operations as Lorentz transformations, thereby avoiding the reliance on tangent-space approximations. Beyond purely hyperbolic models, \citet{skopek2020mixed} introduced mixed-curvature variational autoencoders, whose latent spaces were composed of multiple constant-curvature manifolds, enabling generative models to benefit from diverse geometric structures simultaneously. More recently, \citet{yang2024hypformer} explored an efficient Transformer architecture in the Lorentz model of hyperbolic space, providing hyperbolic counterparts for essential modules such as positional encodings, layer normalization, and residual connections. Parallel to architectural advances, \citet{yang2024hyperbolic} investigated fine-tuning Euclidean LLMs directly in hyperbolic space, demonstrating improved downstream performance by leveraging the inherent hierarchical structure of token embeddings. Building on this line of work, \citet{he2025helm} introduced Hyperbolic LLMs with a Mixture-of-Curvature Experts design, where each expert resided in a hyperbolic manifold of distinct curvature, allowing flexible encoding of input sequences and showcasing the scalability of curvature-aware modeling in large-scale pretraining. These methods still use single geometry for LLMs while we aim to go beyond that to multiple geometries simultaneously.

\section{Preliminary}

\subsection{Mixture of LoRA Experts}
Mixture of LoRA Experts consists of a group of $N$ uniform experts $\{E_i\}_{i=1}^N$, where each features a parameter-efficient LoRA module to store updated parameters while fine-tuning. Each expert $E_i$ has the following forward process:
\begin{align}
    E_i = B_iA_iX,
\end{align}
where $X \in \mathbb{R}^{d_{\text{in}}\times d}$ denotes the input, tunable weight matrices $A_i\in \mathbb{R}^{r\times d_{\text{in}}}$, $B_i\in \mathbb{R}^{d_{\text{out}}\times r}$, and $r \ll \min\{d_{\text{in}}, d_{\text{out}}\}$ is the maximum rank attainable by the trainable matrix, with matrix $A_i$ randomly initialized and matrix $B_i$ set to all zero. The forward process can be formulated as follows:
\begin{align*}
   O  = WX + \sum_{i=1}^{N} R(X)_i E_i = WX + \sum_{i=1}^{N} R(X)_i B_iA_iX,
\end{align*}
where $W$ is the frozen pretrained weight of the feed-forward neural network (FFN) block, $R(X)$ is the token router in the MoE module which routes each token into several distinct experts (e.g., top-$K$ experts for sparse MoE \citep{fedus2022switch}) among all $N$ experts. 



\subsection{Lorentz Model of Hyperbolic Space}
\textbf{Lorentz Model.} The Lorentz model, also called the hyperboloid model, provides one of the isometric realizations of the hyperbolic space as a Riemannian manifold. Formally, an $n$-dimensional Lorentz model with constant negative curvature $\kappa < 0$ is defined as:
\[
\mathbb{L}^{n,\kappa} = \left\{ \mathbf{x} \in \mathbb{R}^{n+1} \;\middle|\; \langle \mathbf{x}, \mathbf{x} \rangle_{\mathcal{L}} = {1}/{\kappa},\; x_t > 0 \right\},
\]
where $\mathbf{x} = [x_t;\, \mathbf{x}_s]^\top$ with $x_t \in \mathbb{R}$ and $\mathbf{x}_s \in \mathbb{R}^n$, and the Lorentzian inner product is given by:
\[
\langle \mathbf{x}, \mathbf{y} \rangle_{\mathcal{L}} = -x_t y_t + \mathbf{x}_s^\top \mathbf{y}_s 
= \mathbf{x}^\top \mathrm{diag}(-1, 1, \dots, 1)\,\mathbf{y}.
\]
Geometrically, $\mathbb{L}^{n,\kappa}$ corresponds to the upper sheet of a two-sheeted hyperboloid embedded in $(n+1)$-dimensional Minkowski space, with the distinguished coordinate $x_t$ representing the time-like axis and the remaining $n$ coordinates forming the space-like axes. This construction not only aligns with the terminology of special relativity \citep{resnick1991introduction} but ensures numerical stability in optimization tasks.

\begin{figure*}[!t]
\begin{center}
\includegraphics[width=0.83\linewidth]{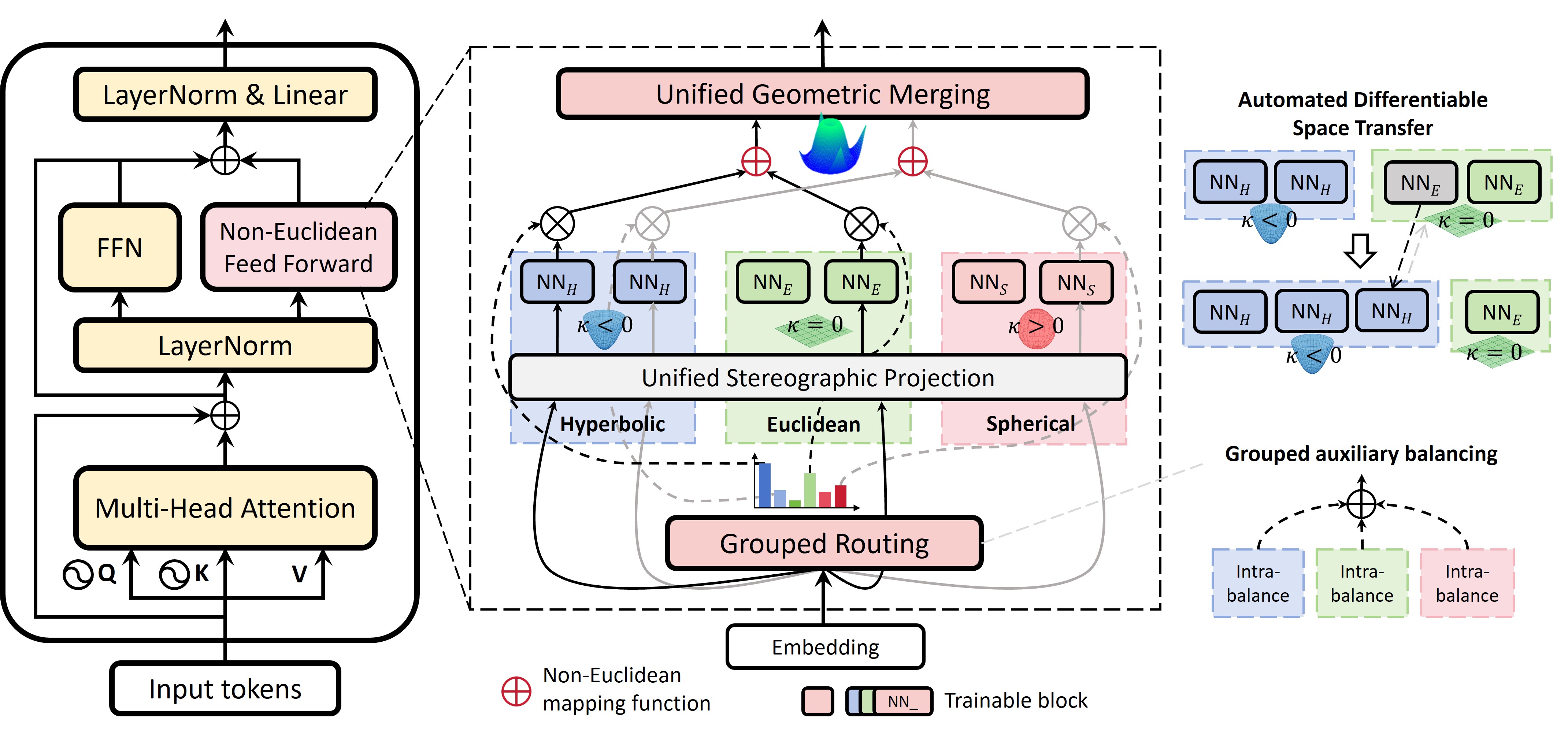}
\end{center}
\caption{The MoSLoRA architecture contains heterogeneous geometric experts unified in our MoS scheme, with various curvatures. Three geometric expert groups are embedded into the FFN layer and labeled in different colored blocks. The grouped auxiliary balancing enforces balanced routing within each space group while allowing free inter-space transitions, and remains fully reversible and differentiable.}
\label{fig:arch}
\end{figure*}

\paragraph{Tangent Space and Maps.} 
For each point $\mathbf{x} \in \mathbb{L}^{n,\kappa}$, tangent space $\mathcal{T}_{\mathbf{x}}\mathbb{L}^{n,\kappa}$ is defined as the Lorentz-orthogonal complement of $\mathbf{x}$ and constitutes a smooth Euclidean subspace of $\mathbb{R}^{n+1}$. Tangent space provides a local linear approximation of the curved manifold for optimization and representation learning. The transition between the manifold and its tangent space is realized through exponential and logarithmic maps. The exponential map $\exp_{\mathbf{x}}^{\kappa}: \mathcal{T}_{\mathbf{x}}\mathbb{L}^{n,\kappa} \to \mathbb{L}^{n,\kappa}$ takes a tangent vector $\mathbf{u} \in \mathcal{T}_{\mathbf{x}}\mathbb{L}^{n,\kappa}$ and projects it onto the manifold along the geodesic starting at $\mathbf{x}$, which is: 
\begin{equation}
\label{expmap}
\exp_{\mathbf{x}}^{\kappa}(\mathbf{u}) 
= \cosh\!\left(\sqrt{|\kappa|}\|\mathbf{u}\|_{\mathcal{L}}\right)\mathbf{x} 
+ \frac{\sinh\!\left(\sqrt{|\kappa|}\|\mathbf{u}\|_{\mathcal{L}}\right)}{\sqrt{|\kappa|}\|\mathbf{u}\|_{\mathcal{L}}}\,\mathbf{u}.
\end{equation}
Conversely, the logarithmic map $\log_{\mathbf{x}}^{\kappa}: \mathbb{L}^{n,\kappa} \to \mathcal{T}_{\mathbf{x}}\mathbb{L}^{n,\kappa}$ takes a point $\mathbf{y} \in \mathbb{L}^{n,\kappa}$ and returns the unique tangent vector at $\mathbf{x}$ that corresponds to the geodesic connecting $\mathbf{x}$ and $\mathbf{y}$:
\begin{equation}
\log_{\mathbf{x}}^{\kappa}(\mathbf{y}) 
= \frac{\cosh^{-1}\!\big(\kappa \langle \mathbf{x}, \mathbf{y}\rangle_{\mathcal{L}}\big)}{\sinh\!\big(\cosh^{-1}(\kappa \langle \mathbf{x}, \mathbf{y}\rangle_{\mathcal{L}})\big)} \left( \mathbf{y} - \kappa \langle \mathbf{x}, \mathbf{y}\rangle_{\mathcal{L}} \mathbf{x} \right).
\end{equation}
These two maps establish a rigorous correspondence between the locally Euclidean tangent space and the globally curved hyperbolic manifold.

\section{Mixture
of Space Experts Framework}
\label{Method}


In this section, we propose our Mixture of Space tuning scheme that adapts the model to the implicit curvature of semantic subspaces while autonomously exploring and transitioning between different constant curvatures within the same subspace, and also across distinct subspaces. 


\subsection{Mixture of Space Scheme}

We explore three constant-curvature spaces with positive, negative, and zero curvature, commonly modeled as spherical $\mathbb{S}$, hyperbolic $\mathbb{H}$, and Euclidean $\mathbb{E}$ spaces. For a $n$-dimensional hyperbolic space $\mathbb{H}^n_{\kappa}$ with constant negative curvature $\kappa$, each point $\mathbf{x}\in\mathbb{R}^{n+1}$ in $\mathbb{H}^n_{\kappa}$ should satisfy:
\begin{equation}
    \mathbb{H}^n_{\kappa} := \{\mathbf{x}\in\mathbb{R}^{n+1} | \left\langle \mathbf{x}, \mathbf{x} \right\rangle_{L}=1/\kappa, \kappa<0\},
\end{equation}
where $\mathbb{H}^n_{\kappa}$ is defined by the Lorentz inner product $\left\langle \mathbf{x}, \mathbf{x} \right\rangle_{L} = -x_1^2+\sum^{n+1}_{i=2}x_i^2$, and $\mathbf{x}=[x_1, x_2,\dots,x_{n+1}]=[x_{\text{time}},\mathbf{x}_{\text{space}}]$ denotes an arbitrary point with time-like component $x_1$ and space-like component $[x_2,\dots,x_{n+1}]$. In Lorentz model of hyperbolic space, the volume $V_{\kappa}$ grows exponentially with its radius $r$ as $V_{\kappa}(r) \asymp \exp\left((n-1)\sqrt{-\kappa}\, r\right)$. 
Thus, the larger the magnitude of $\kappa$, the greater the curvature and the faster the volume expansion, allowing the space to accommodate more hierarchical structures with higher representational capacity.
For spherical spaces with curvature $\kappa >0$, each point should satisfy:
\begin{align}
    \mathbb{S}^n_{\kappa} := \{\mathbf{x}\in\mathbb{R}^{n+1} | \left\langle \mathbf{x}, \mathbf{x} \right\rangle_{2}=1/\kappa, \kappa>0\},
\end{align}
where $\left\langle \cdot, \cdot \right\rangle_{2}$ is the Euclidean inner product and $\left\langle \mathbf{x}, \mathbf{x} \right\rangle_{2} = \sum^{n+1}_{i=1}x_i^2$. Unlike hyperbolic spaces, which are suitable for representing complex hierarchical structures, volume in spherical spaces grows more slowly than Euclidean space with respect to radius, hence they are especially suitable for modeling cyclic, periodic, or bounded structures, where global capacity is limited but dense local clustering and angular relationships (e.g., directions, orientations) are crucial. 

To unify the three aforementioned spaces and associated constraints, we introduce a differentiable scheme, \textbf{Mixture of Space} (\textbf{MoS}), which preserves the local degrees of freedom and the flexibility of transformations across all three spaces. The formal definition is given as follows:
\begin{equation}
\text{G}(\mathbf{x}) = \left[\begin{array}{c}
\sqrt{\|\mathbf{s}\|^2 \cdot \mathbf{sgn}(-\kappa) + \varphi(\kappa)} \\
\mathbf{s}
\end{array}\right] = 
\left[\begin{array}{c}
\xi^{\prime} \\ \mathbf{s}
\end{array}\right]
\label{unified_scheme}
\end{equation}
\begin{equation*}
\mathbf{sgn}(\kappa) :=
\begin{cases} 
-1, & \kappa < 0, \\
1, & \kappa \geq 0,
\end{cases}
\quad
\varphi(\kappa) :=
\begin{cases} 
1/\lvert \kappa \rvert, & \kappa \neq 0, \\
0, & \kappa = 0.
\end{cases}
\end{equation*}
where point $\mathbf{x}=(\xi; \mathbf{s}^T)^T\in\mathbb{R}^{n+1}$, $\xi,\xi^{\prime}\in\mathbb{R}, \mathbf{s}\in\mathbb{R}^{n}$, and $\text{G}(\mathbf{x})$ denotes the unified space transformation from different geometries with different signs of the curvature $\kappa$. We restrict transformation to the space-like component, as Lorentz geometry in
$\mathbb{R}^{n+1}$ provides only $n$ degrees of freedom, yielding better numerical stability. While maintaining the original space definitions, we further extend vectors in the Euclidean space with an additional dimension, enabling seamless coupling among the three spaces. This differentiable scheme enables embeddings to be jointly learned across manifolds, capturing complementary structural information and maintaining geometric consistency. 
\subsection{Unified Stereographic Projection}
While existing mapping schemes are capable of strictly preserving hierarchical structures \citep{chen2021fully,bdeir2023fully,yang2024hyperbolic}, they depend on computationally expensive and GPU-unfriendly operations, such as exponential and logarithmic maps in Eq.\eqref{expmap}, and therefore lack a unified, end-to-end differentiable formulation. To address this challenge, we propose a unified stereographic projection, as illustrated in the middle of Fig.\ref{fig:arch}, which projects inputs from the input space into three constant-curvature spaces and subsequently maps them back to the output space. In our unified stereographic projection, 
each point $\mathbf{x}$ (from input space) will be firstly transmitted into a projected space through stereographic conformal inverse projection $\rho_{\kappa}^{-1}(\cdot)$ in Eq.\eqref{invproj} to $(\xi; \mathbf{s}^T)^T\in\mathbb{R}^{n+1}$, where ${\kappa} \in \mathbb{R}$ is the curvature of the projected embedding space:
\begin{align}
\label{invproj}
&\rho_{\kappa}^{-1}(\mathbf{x}) = \left( \frac{1}{\sqrt{|\kappa|}} \frac{1 - \kappa \|\mathbf{x}\|_2^2}{1 + \kappa \|\mathbf{x}\|_2^2}, \frac{2\mathbf{x}}{1 + \kappa \|\mathbf{x}\|_2^2} \right)^T = 
\left[\begin{array}{c}
\xi \\ \mathbf{s}
\end{array}\right]
\end{align}
Then, after model-related transformation from $(\xi; \mathbf{s}^T)^T$ into $(\xi^{\prime}; {\mathbf{s}^{\prime}}^T)^T$, tokens will be projected back to the original space by stereographic projection:
\begin{align}
\label{proj}
\rho_{\kappa}((\xi^{\prime}; {\mathbf{s}^{\prime}}^T)^T) = \mathbf{s}^{\prime}/({1 + \sqrt{|\kappa|} \xi^{\prime}}),
\end{align}
To guarantee constraint satisfaction and numerical stability,
we apply a scaling factor prior to the inverse projection and
rescale the output afterward.
As formalized in Lemma~\ref{lemma1}, this procedure is theoretically equivalent to the
original formulation, with a detailed proof provided in
Appendix~\ref{appdix:equivar}.
\begin{lemma}[Scale-Invariant Equivalence of the Unified Projection]
\label{lemma1}
Let $\kappa \neq 0$ and let $c>0$ be a variable scaling constant.
Consider the unified projection and MoS transformation as $F_{\kappa}(\cdot)$ defined in
Eq.\eqref{unified_scheme}, \eqref{invproj}, and \eqref{proj}.
Applying a scaling factor $\gamma$ to the input before the inverse projection
and rescaling the output by the same factor yields:
\[
F_{\kappa}(x)
\;=\;
1/{\gamma}\cdot F_{\kappa / \gamma ^{2}}(\gamma x).
\]
\end{lemma}
To that end, our unified stereographic projection enables consistent and efficient bidirectional transformations across all subspaces within a single cohesive architecture.

\subsection{Mixture of Space Parameters-Efficient Tuning}



Building upon the unified geometric formulation introduced earlier in Eq.\eqref{unified_scheme}, we now specialize it to a mixture-of-experts PEFT setting and introduce our \textbf{MoSLoRA}. Specifically, given a projected embedding $(\xi_i; \mathbf{s}_i^T)^T \in \mathbb{R}^{n+1}$ of token $\mathbf{x}_i$, we treat the space-like component $\mathbf{s}_i \in \mathbb{R}^n$ as the routing unit for MoSLoRA within each layer’s FFN block $W$, as is illustrated in Figure \ref{fig:arch}. Each space expert is defined as:
\begin{equation*}
\text{G}(W,\mathbf{x}_i) = \left[\begin{array}{c}
\sqrt{\|W \mathbf{s}_i\|^2 \cdot \mathbf{sgn}(-\kappa) + \varphi(\kappa)} \\
W \mathbf{s}_i
\end{array}\right] = 
\left[\begin{array}{c}
\xi^{\prime}_i \\ \mathbf{s}^{\prime}_i
\end{array}\right]
\end{equation*}

The output tokens $(\xi^{\prime}_i; {\mathbf{s}^{\prime}_i}^T)^T \in \mathbb{R}^{n+1}$ from the top-$K$ geometric experts, each operating on distinct curved spaces and capturing various relations, are projected back into Euclidean space, where these relations are preserved and merged together. The non-Euclidean feed-forward output is formulated as follows: 
\begin{align}
\mathbf{o}_i = \sum^N_{i=1}\sum^Q_{j=1} R\left(\mathbf{x}_j\right)_i \rho_{\kappa_i}((\xi_j; {\mathbf{s}^{\prime}}^T_j)^T),
\end{align}
where $\rho_{\kappa_i}((\cdot;\cdot))$ is the stereographic projection to map points (token embeddings) from the curved space back to the output space, $\kappa_i$ is the curvature of the $i$-th geometric expert, $R(x_j)_i$ is the routing value for token $x_j$ and $i$-th expert. On top of this, we propose a new \textbf{grouped auxiliary loss} (in Fig.~\ref{fig:arch}) to balance top-$K$ routing within each heterogeneous geometric expert group to encourage balancing among experts in the same space instead of among different spaces.
It is worth noting that the curvature $\kappa$ of each expert is a learnable parameter, allowing the model to independently adjust the sub-space of each expert to approach the optimal geometry. Additionally, from an efficiency perspective, the stereographic mapping and the associated routing strategy used here require neither complex exponential/logarithmic space operations nor prior knowledge of the latent embedding space, thereby yielding higher efficiency in both training and inference. We formally prove that our MoSLoRA framework admits uniformly bounded gradients across different geometric curvatures, ensuring training stability; detailed proofs are deferred to the Appendix \ref{appendix:grad_bound}.

\textbf{Separated Optimizers.} Considering that curvature, as a geometric parameter, differs fundamentally from other capacity-related model weights, we assign it an independent optimizer to encourage further exploration of the curvature of the latent space while reducing the risk of overfitting to local data patterns. 
Formally, $\Theta = \{\kappa^{(1)}, \dots, \kappa^{(K)}, \;\theta^{(1)}, \dots, \theta^{(M)}\}$ consists of curvature parameters $\kappa^{(j)}$ and capacity-related parameters $\theta^{(m)}$. To that end, we have:
\begin{align}
    w \;\leftarrow\; 
w &- \sum_{j=1}^{N} \eta_\kappa^{(j)} \,\mathbf{1}_{w \in \kappa^{(j)}} \cdot g_\kappa^{(j)}(w)\nonumber \\
   &- \sum_{m=1}^{M} \eta_\theta^{(m)} \,\mathbf{1}_{w \in \theta^{(m)}} \cdot g_\theta^{(m)}(w),
\end{align}
where $\eta_\kappa^{(j)}$ and $\eta_\theta^{(m)}$ denote their respective learning rates, 
$g_\kappa^{(j)}(\cdot)$ and $g_\theta^{(m)}(\cdot)$ are the corresponding gradients, 
and $\mathbf{1}_{(\cdot)}$ is an indicator function selecting the parameter group.

\begin{table*}[!t]
\centering
\caption{Performance comparison of different tuning methods across different natural language understanding and mathematical reasoning benchmarks. Avg. column reports the average score, and $\text{rank}\uparrow$ indicates that a larger rank parameter is used compared to other methods.
}
\label{tb:main_result}
\begin{small}
\begin{tabular}{l|ccccccc|c}
\toprule
\textbf{Methods} & \textbf{COLA} & \textbf{MRPC} & \textbf{GSM8k} & \textbf{MATH500} & \textbf{MAWPS} & \textbf{SVAMP} & \textbf{AQuA} & \textbf{Avg.} \\
\midrule
\textbf{Base} (Qwen2-1.5B) & 32.41 & 69.16 & 33.13 & 0.0 & 43.85 & 54.33 & 31.10 & 37.71 \\
\textbf{LoRA} & 87.25 & 87.30 & 44.58 & 15.60 & 63.85 & 63.33 & 28.35 & 55.75 \\
\textbf{DoRA} & 87.15 & 87.19 & 42.53 & 14.20 & 62.12 & 62.33 & 30.71 & 55.18 \\
\textbf{LoRA} $\text{rank}\uparrow$ & 83.51 & 86.43 & 46.85 & 13.60 & 61.92 & 60.33 & \textbf{33.46} & 55.16 \\
\textbf{AdaLoRA} & 82.45 & 79.77 & \textbf{51.33} & 16.60 & 62.50 & 58.67 & 23.23 & 53.51 \\
\textbf{MELoRA} & 86.96 & 86.55 & 44.96 & 16.40 & 62.69 & \textbf{65.67} & 32.28 & 56.50 \\
\textbf{HMoRA} & 66.63 & 66.49 & 12.28 & 2.60 & 54.23 & 46.00 & 24.41 & 38.95 \\
\textbf{HypLoRA} & 85.91 & 87.71 & 45.56	& 14.60	& 62.88	& 64.67	& 27.95 & 55.61\\
\textbf{HydraLoRA} & \underline{87.54} & \underline{88.17} & 43.67 & 16.80 & 62.88 & 62.67 & 28.35 & 55.73 \\
\midrule
\rowcolor{lightgray!29} \textbf{MoSLoRA (Ours)} & \textbf{87.63} & \textbf{88.23} & \underline{47.76} & \textbf{18.00} & \textbf{74.62} & \underline{64.33} & \underline{31.50} & \textbf{58.87} \\
\bottomrule
\end{tabular}
\end{small}
\end{table*}

\begin{figure*}[!t]
    \centering
    \begin{subfigure}{0.24\textwidth}
        \includegraphics[width=\linewidth]{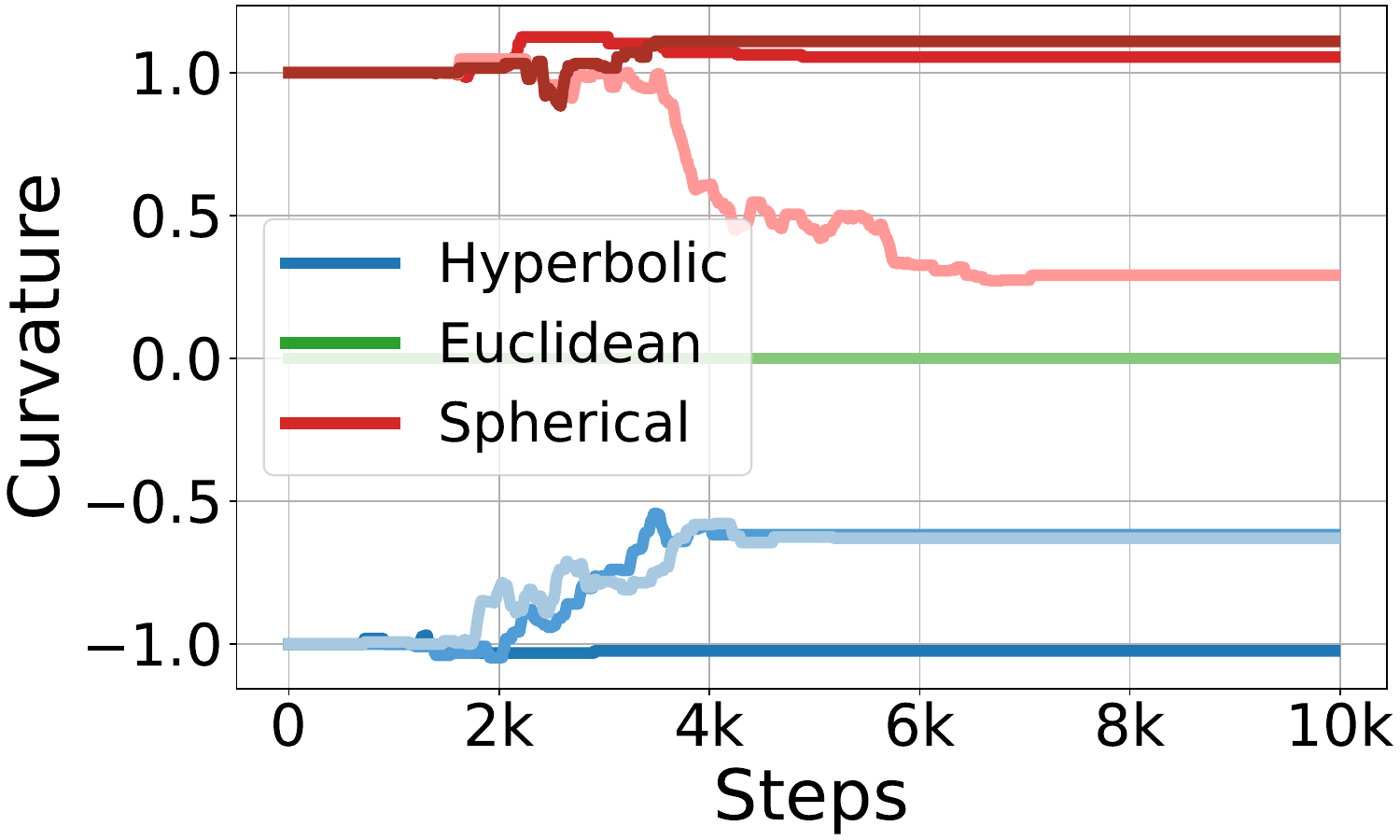}
        \caption{Layer-1}
        \label{fig:image1}
    \end{subfigure}
    \hfill
    \begin{subfigure}{0.24\textwidth}
        \includegraphics[width=\linewidth]{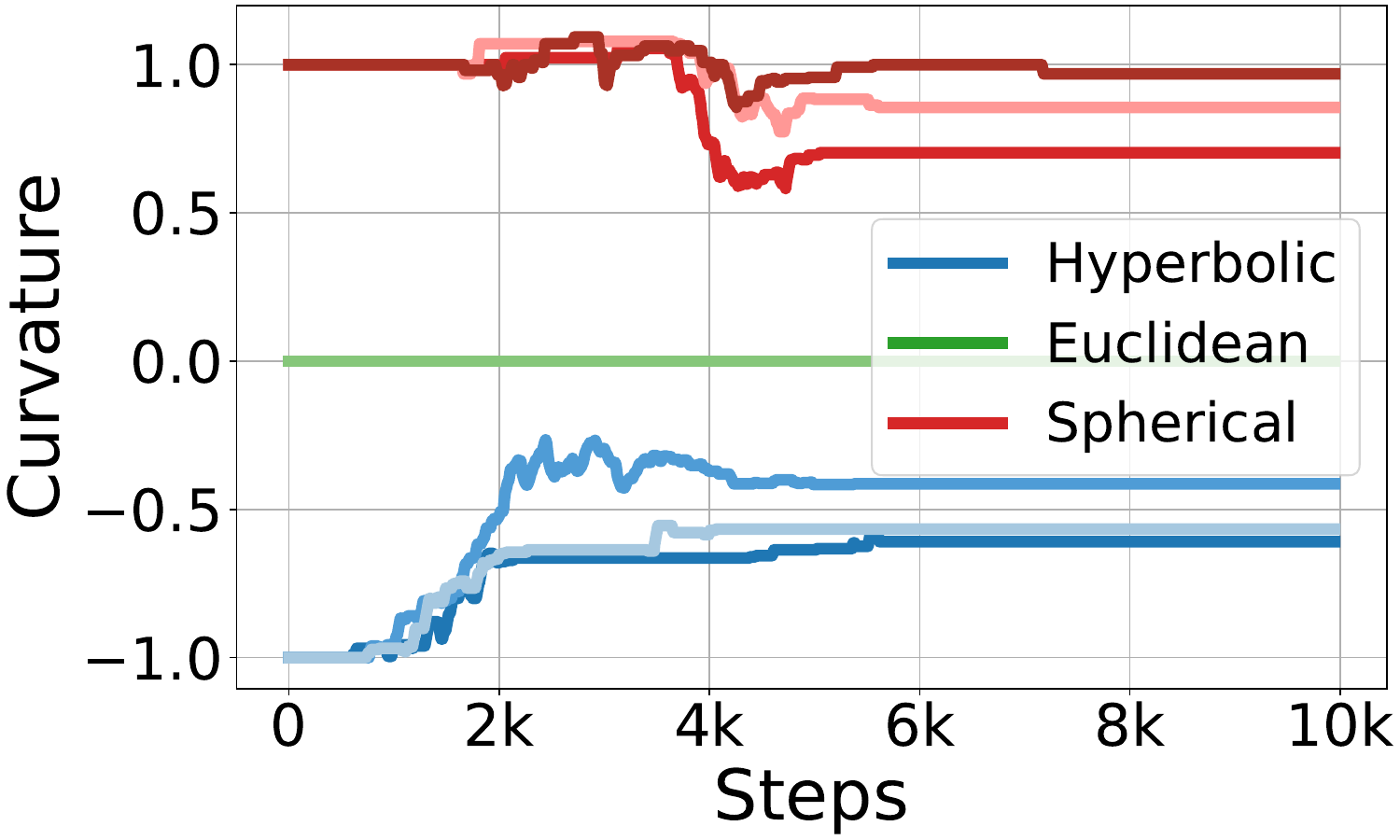}
        \caption{Layer-10}
        \label{fig:image2}
    \end{subfigure}
    \hfill
    \begin{subfigure}{0.24\textwidth}
        \includegraphics[width=\linewidth]{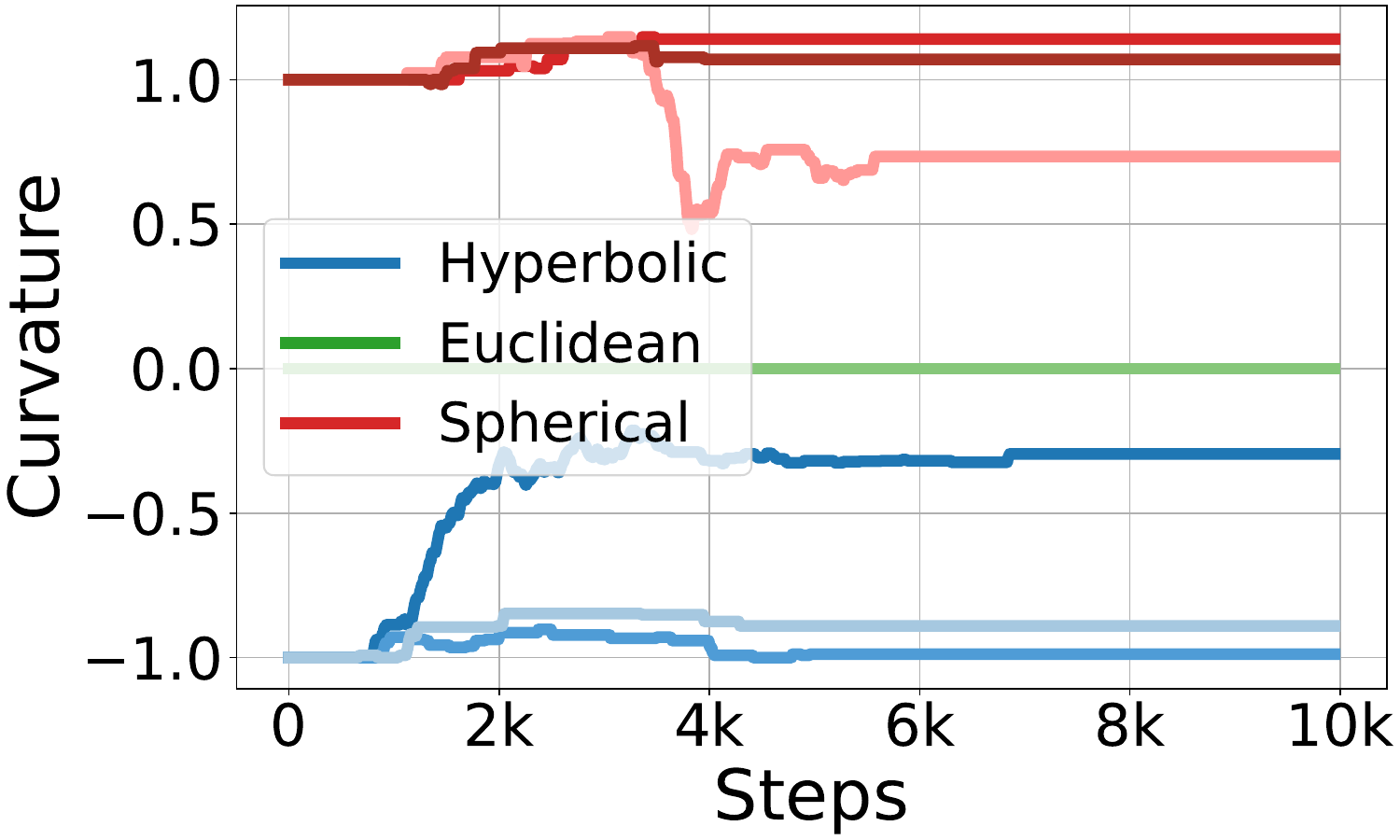}
        \caption{Layer-15}
        \label{fig:image3}
    \end{subfigure}
    \hfill
    \begin{subfigure}{0.24\textwidth}
        \includegraphics[width=\linewidth]{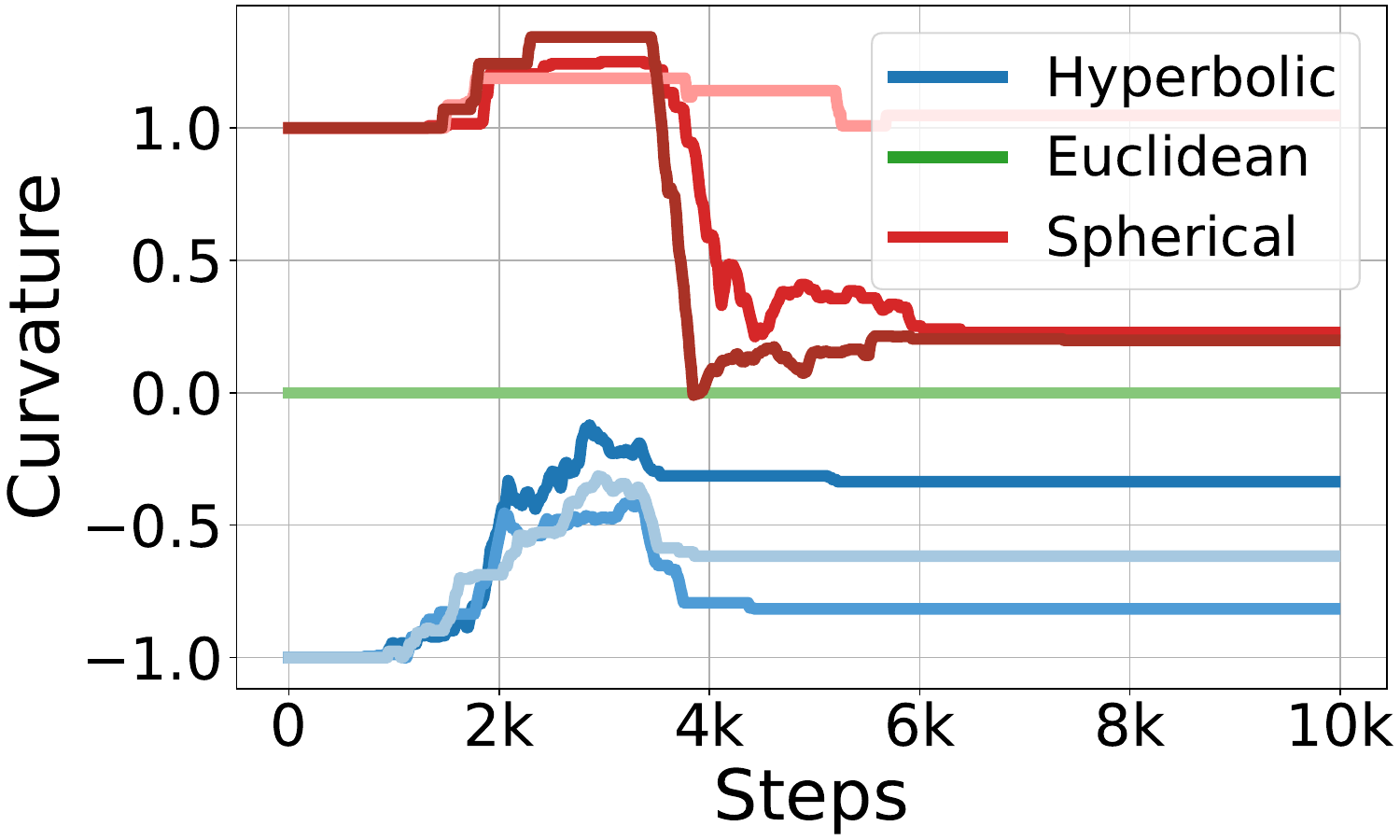}
        \caption{Layer-20}
        \label{fig:image4}
    \end{subfigure}
    \caption{Curvature dynamics of each geometric expert of different layers during training.}
    \label{fig:curvature_dynamic_layer}
\end{figure*}

\section{Experiments}
\label{experiments}

\subsection{Experiment Setting}

\textbf{Dataset and Benchmarks.}
To explore the utility of our method, we evaluate our approach on both natural language understanding \citep{wang2018glue} and mathematical reasoning datasets. For NLP tasks, we adopt the Microsoft Research Paraphrase Corpus (MRPC) \citep{dolan2005automatically}, Recognizing Textual Entailment (RTE) \citep{bowman-etal-2015-large} and the Corpus of Linguistic Acceptability (CoLA) \citep{warstadt-etal-2019-neural}. For mathematical reasoning, we employ GSM8K \citep{cobbe2021training}, MAWPS \citep{koncel2016mawps}, SVAMP \citep{patel2021nlp}, and AQuA \citep{DBLP:journals/corr/LingYDB17}, alongside the MATH500 by OpenAI \citep{lightman2023let}, a curated challenging subset of the MATH benchmark \citep{hendrycks2021measuring}. For commonsense resaonsing, we include OpenBookQA (OBQA)~\citep{mihaylov-etal-2018-suit} and CommonsenseQA (CSQA)~\citep{talmor-etal-2019-commonsenseqa}. The training set is constructed by uniformly sampling from the mathematical datasets except MATH500, which is reserved for evaluating zero-shot and few-shot performance on unseen problems. 



\textbf{Base Model and Baselines.}
We include Qwen2-1.5B \citep{team2024qwen2}, Qwen2.5-3B \citep{qwen2025qwen25technicalreport}, and Gemma2-2B-it \citep{team2024gemma}
as our base models. 
For parameter-efficient fine-tuning methods, we compare our method with LoRA \citep{hu2022lora}, AdaLoRA \citep{zhang2023adalora}, DoRA \citep{liu2024dora}, MELoRA \citep{ren2024melora}, HMoRA \citep{liao2025hmora}, HydraLoRA \citep{tian2024hydralora} and HypLoRA \citep{yang2024hyperbolic}. 
These methods share the same training settings as ours, and some also incorporate architectures related to the MoE design. 
In this work, we adopt an asymmetric architecture similar to \citet{tian2024hydralora}, minimizing the activated parameters while verifying the effectiveness of our proposed scheme.

\begin{table*}[t]
\centering
\caption{Performance comparison on math and commonsense reasoning, and language understanding benchmarks based on Qwen2.5-3B.}
\begin{small}
\begin{tabular}{l|cccccccccc|c}
\toprule
\textbf{Methods} & \textbf{GSM8k} & \textbf{MATH500} & \textbf{MAWPS} & \textbf{SVAMP} & \textbf{AQuA} & \textbf{MRPC} & \textbf{CoLA} & \textbf{RTE} & \textbf{CSQA} & \textbf{OBQA} & \textbf{Avg.} \\
\midrule
Base                 & 59.06 & 0.40  & 60.19 & 74.67 & 50.00 & 56.41 & 62.80 & 51.62 & 32.40 & 32.40 & 47.99 \\
LoRA    & \textbf{65.50} & 25.60 & 65.58 & 76.33 & 38.98 & 87.71 & 69.13 & 89.53 & 82.80 & 87.00 & 68.81 \\
HydraLoRA   & 64.52 & 22.40 & 65.77 & 75.33 & \textbf{42.91} & \textbf{88.46} & 85.91 & 89.89 & \textbf{82.56} & \textbf{89.20} & 70.69 \\
HypLoRA   & 64.67 & 26.40 & 65.77 & \textbf{78.67} & 41.34 & 88.23 & \textbf{87.34} & 90.25 & 79.85 & 87.80 & 71.03 \\
\textbf{MoSLoRA}  & \underline{\textbf{63.68}} & \textbf{29.60} & \textbf{81.73} & \underline{\textbf{78.00}} & \underline{\textbf{39.37}} & \underline{\textbf{88.41}} & \underline{\textbf{86.19}} & \textbf{90.97} & \underline{\textbf{82.47}} & \underline{\textbf{89.00}} & \textbf{72.94} \\
\bottomrule
\end{tabular}
\end{small}
\label{tab:main_results-qwen2.5-3B}
\end{table*}

\begin{table*}[!t]
\centering
\caption{Performance of different optimizer choices across different MATH and general reasoning benchmarks. \textsc{Uni} and \textsc{Sep} denote using a unified optimizer and separated optimizers, respectively.}
\label{table:optim_split}
\begin{small}
\begin{tabular}{l|ccccccc|c}
\toprule
\textbf{Methods} & \textbf{COLA} & \textbf{MRPC} & \textbf{GSM8k} & \textbf{MATH500} & \textbf{MAWPS} & \textbf{SVAMP} & \textbf{AQuA} & \textbf{Avg.} \\
\midrule
\textbf{MoSLoRA $\textsc{uni}$} & 87.15 & 87.19 & 46.85 & 17.80 & 74.62 & 64.33 & 29.13 & 58.15 \\
\textbf{MoSLoRA $\textsc{sep}$} & \textbf{87.63} & \textbf{88.23} & \textbf{47.76} & \textbf{18.00} & \textbf{74.62} & \textbf{64.33} & \textbf{31.50} & \textbf{58.87} \\
\bottomrule
\end{tabular}
\end{small}
\end{table*}

\begin{figure*}[!t]
    \centering
    \begin{subfigure}{0.23\textwidth}
        \includegraphics[width=\linewidth]{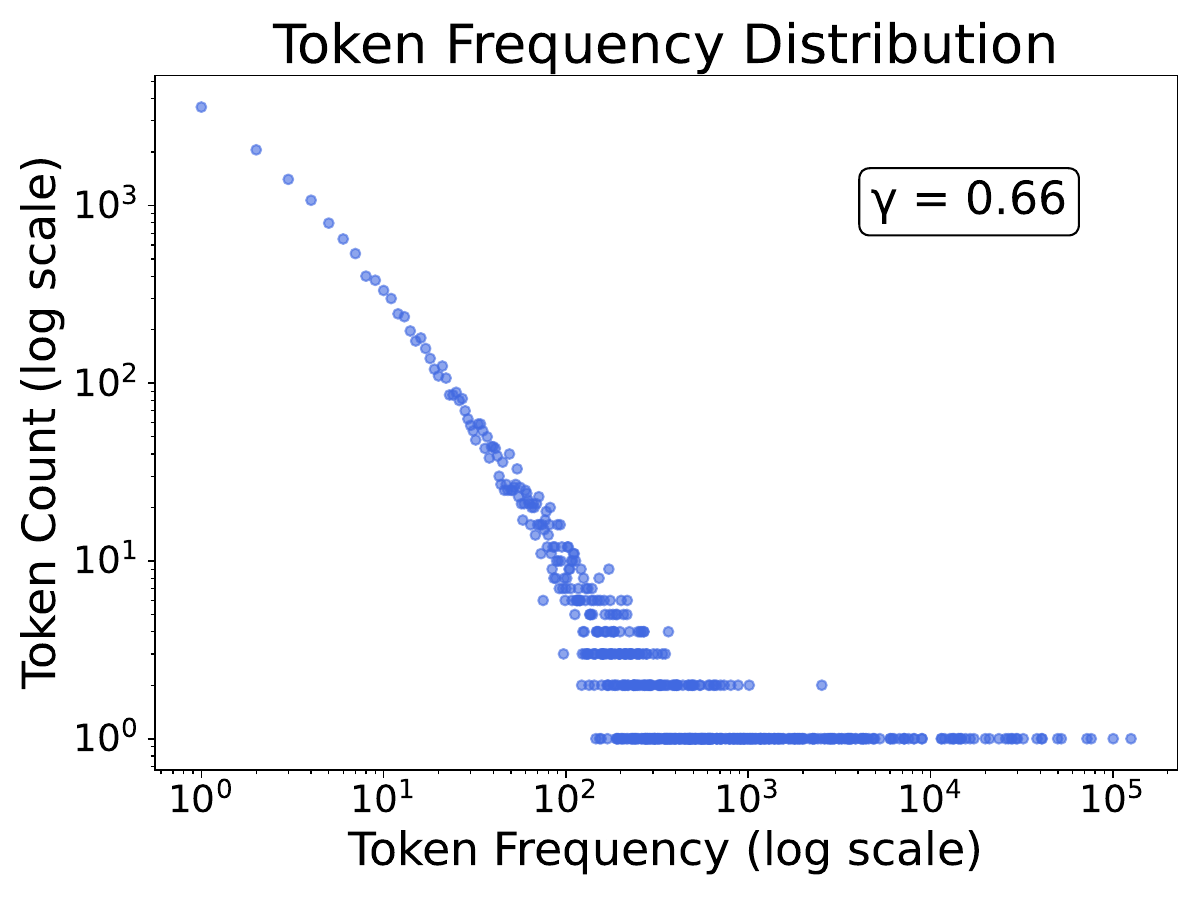}
        \caption{MATH Token Freq.}
        \label{fig:token_image1}
    \end{subfigure}
    \hfill
    \begin{subfigure}{0.23\textwidth}
        \includegraphics[width=\linewidth]{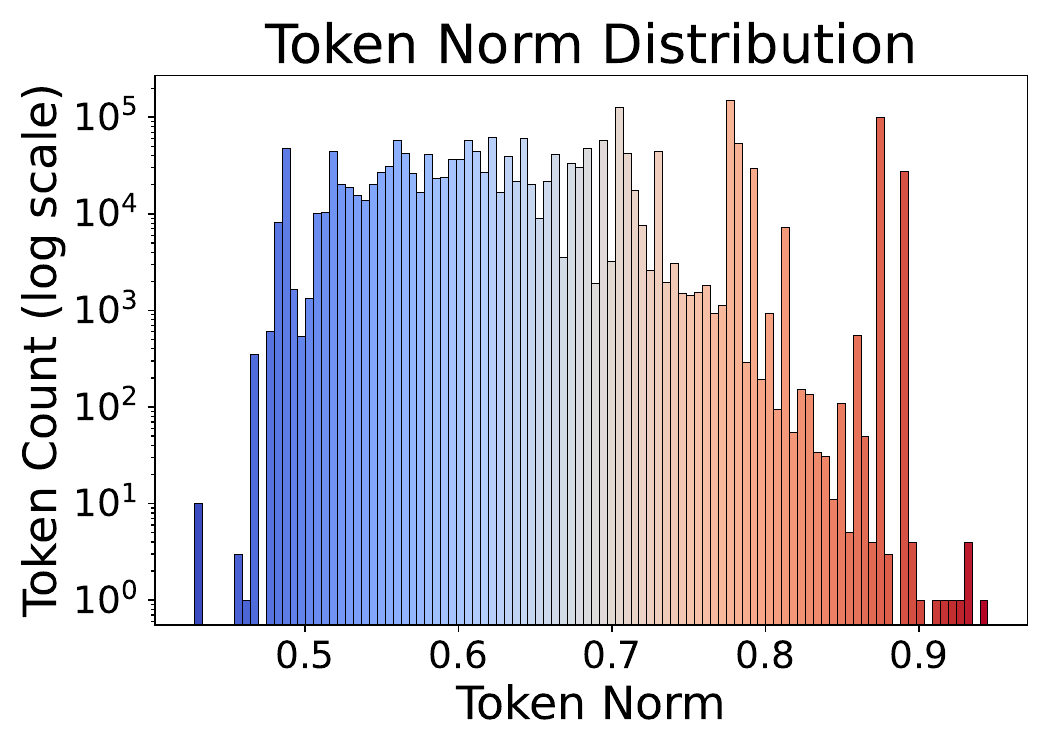}
        \caption{MATH Token Norm}
        \label{fig:token_image2}
    \end{subfigure}
    \hfill
    \begin{subfigure}{0.23\textwidth}
        \includegraphics[width=\linewidth]{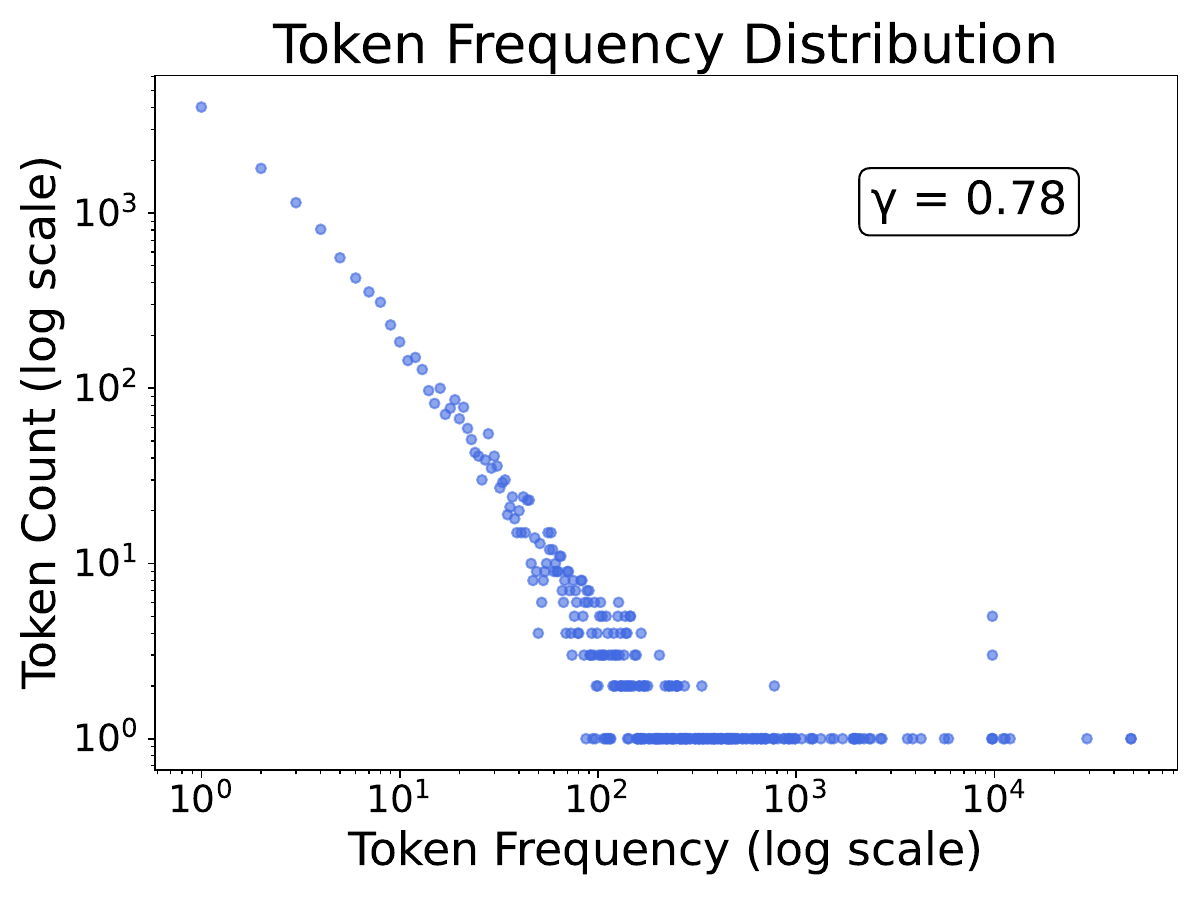}
        \caption{CSQA Token Freq.}
        \label{fig:token_image3}
    \end{subfigure}
    \hfill
    \begin{subfigure}{0.23\textwidth}
        \includegraphics[width=\linewidth]{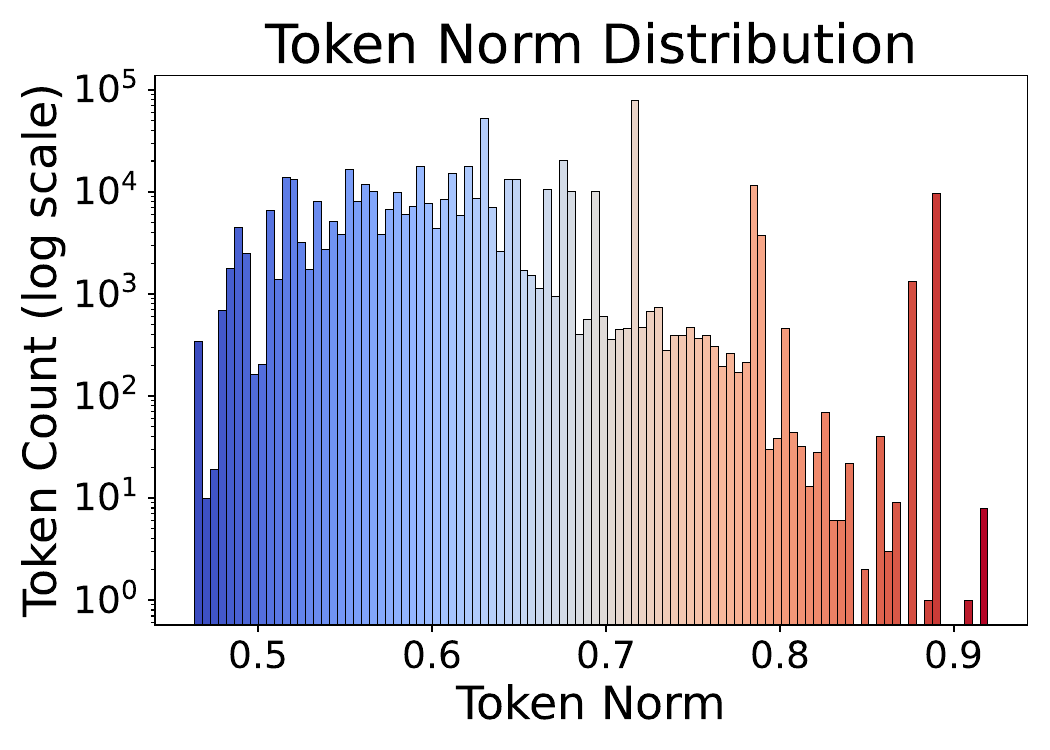}
        \caption{CSQA Token Norm}
        \label{fig:token_image4}
    \end{subfigure}
    \caption{Token norm distribution and token frequency distribution of math reasoning and commonsenseQA datasets.}
    \label{fig:token_freq_norm}
\end{figure*}

\begin{figure}[!t]
    \centering
    \begin{subfigure}{0.48\columnwidth}
        \includegraphics[width=\linewidth]{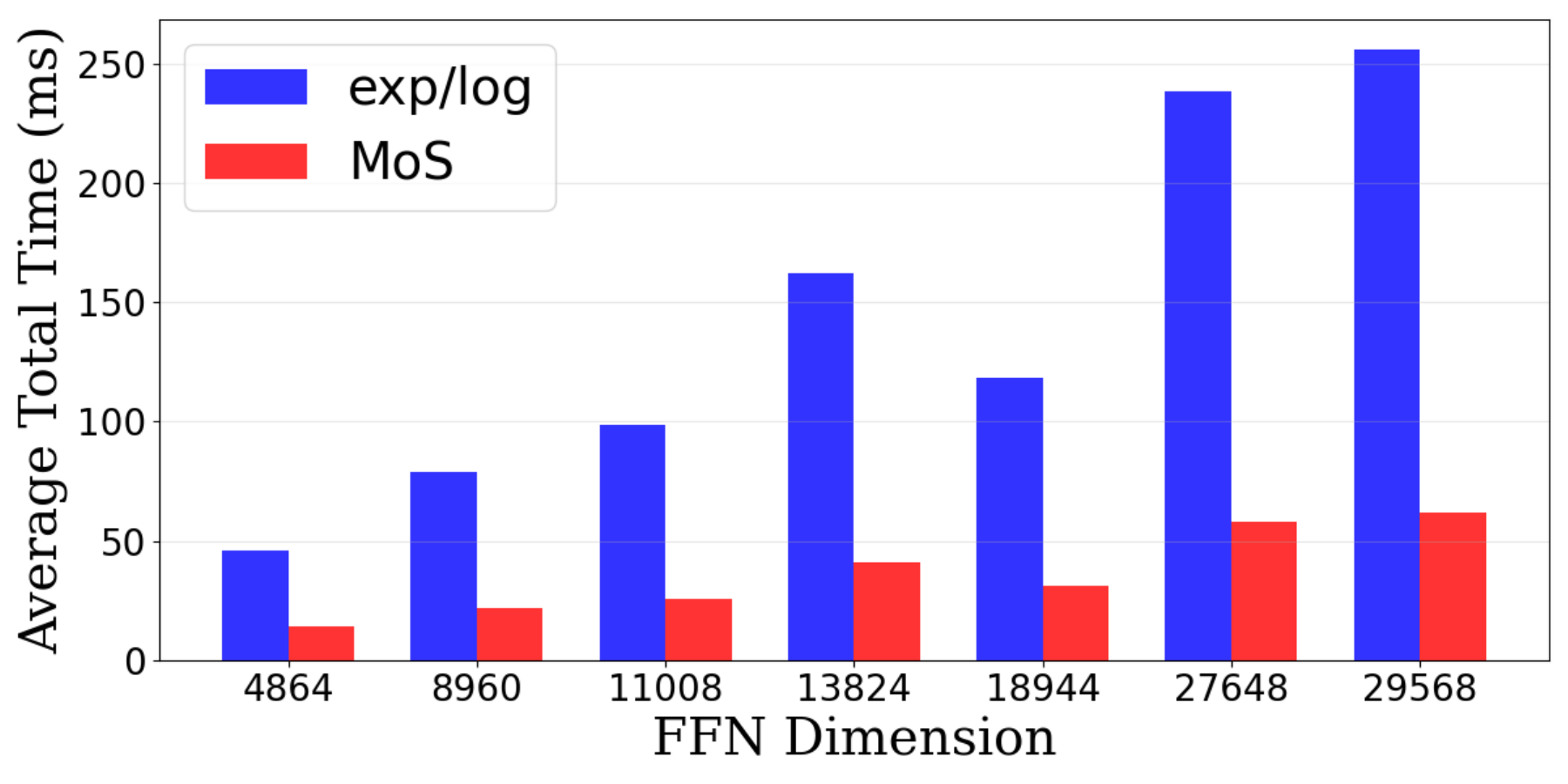}
        \caption{Increasing layer dimensions.}
        \label{fig:image1}
    \end{subfigure}
    \begin{subfigure}{0.48\columnwidth}
        \includegraphics[width=\linewidth]{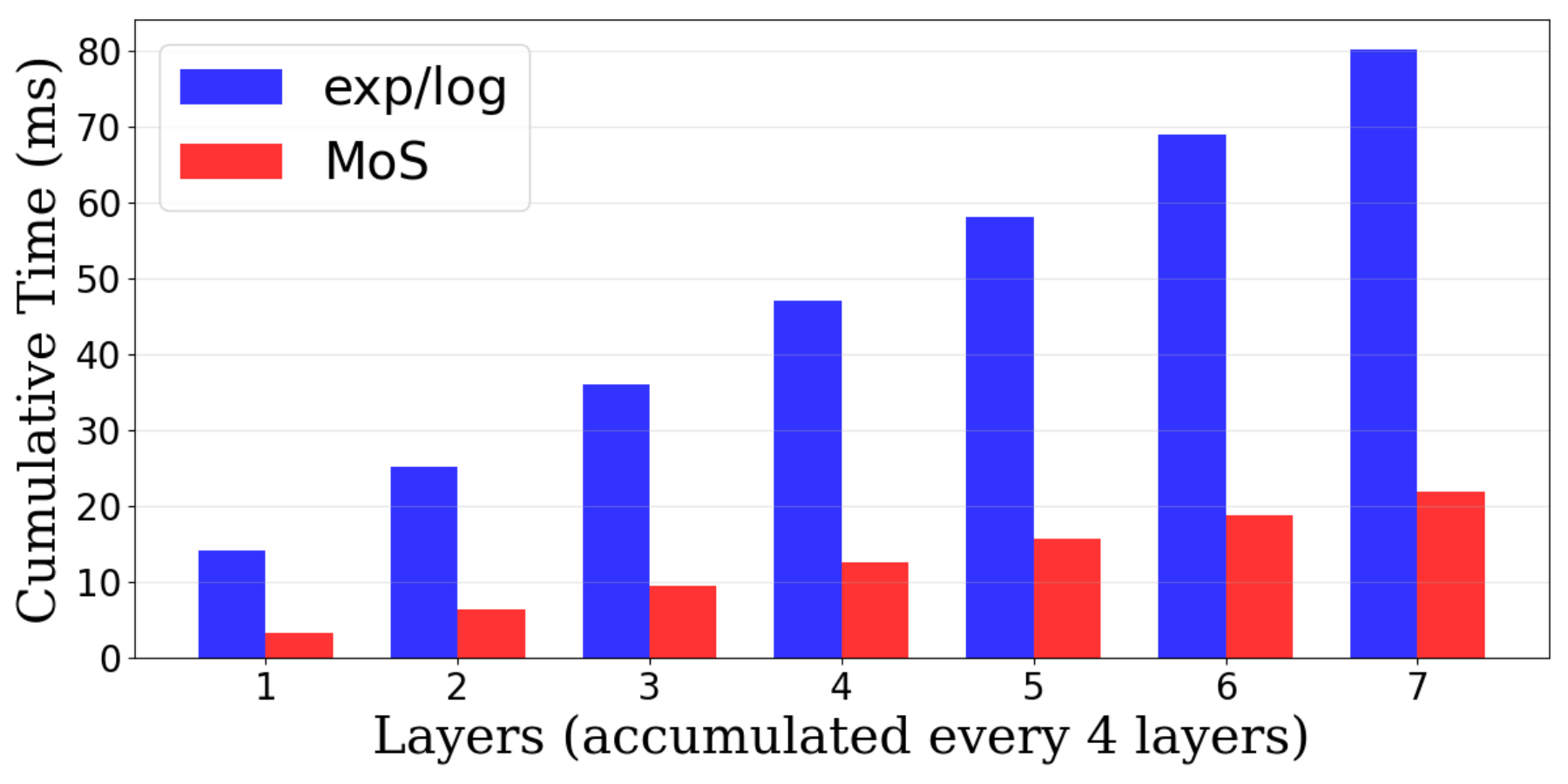}
        \caption{Increasing model layers.}
        \label{fig:image2}
    \end{subfigure}
    \caption{Efficiency comparison between exp/log and our MoS.}
    \label{fig:efficiency_mapping}
\end{figure}
\textbf{Training Settings.}
All models are trained for three epochs on the same dataset using NVIDIA A100 and H800 GPUs. To ensure fairness, we control the number of activated parameters across methods. More detailed information can be found in Appendix~\ref{training_hyperparams}.
\subsection{Baselines Comparison}
We compare our methods with other baselines using 8 distinct experts with top-4 routing. In MoSLoRA, three experts are allocated for each non-Euclidean space group and two for the Euclidean group, with curvatures initialized to –1 for negative curvature and 1 for positive curvature spaces. To keep the number of activated parameters comparable across methods, we vary the LoRA rank from 8 to 64 (0.45\%–3.53\% of activated parameters), while fixing the rank at 8 for all other baselines. Due to their fully activated computation, HMoRA and HydraLoRA yield 2.26\% and 2.54\% activated parameters, respectively, whereas MoSLoRA requires only 1.31\%.

\textbf{Main Results.} As shown in Table \ref{tb:main_result}, our \textbf{MoSLoRA} achieves higher average performance than existing state-of-the-art methods across tasks in both natural language understanding and mathematical reasoning benchmarks. In particular, it obtains the best results on most mathematical reasoning benchmarks, including the challenging MATH500 dataset. Since our training data is uniformly sampled from other mathematical reasoning datasets, these results further demonstrate the superior generalization ability of our \textbf{MoSLoRA} when tackling previously unseen reasoning problems. We observe that our method achieves particularly notable improvements on mathematical reasoning benchmarks compared to existing baselines. We attribute this to the introduction of the mixed-curvature framework, where hyperbolic space is especially well suited for representing hierarchical numerical and logical structures \citep{yang2024hyperbolic}, while spherical space can better capture cyclic properties such as equivalence relations commonly present in mathematics. Consequently, unlike Euclidean baselines that rely solely on a flat embedding space, our approach provides clear advantages in handling large-number computations and symbolic operations. We also scale to 3B model (and other architecture in Appendix~\ref{main_result_gemma2}) and more benchmarks to further validate the effectiveness. As is shown in Table \ref{tab:main_results-qwen2.5-3B}, our \textbf{MoSLoRA} demonstrates strong competitiveness across all evaluated benchmarks, substantially outperforming existing methods on several tasks, and achieving the best overall average performance among all baselines.

\begin{table*}[!t]
\centering
\caption{Performance comparison of geometric-expert mixture recipes across mathematical and general reasoning benchmarks.}
\label{tb:different_mixture}
\begin{small}
\begin{tabular}{l|l|ccccccc|c}
\toprule
\textbf{Methods} & \textbf{Space} & \textbf{COLA} & \textbf{MRPC} & \textbf{GSM8k} & \textbf{MATH500} & \textbf{MAWPS} & \textbf{SVAMP} & \textbf{AQuA} & \textbf{Avg.} \\
\midrule
\textbf{MoSLoRA-S} & Spherical & 86.86 & 87.77 & 44.50 & 17.60 & 72.69 & \textbf{64.33} & 28.35 & 57.44 \\
\textbf{MoSLoRA-H} & Hyperbolic & 85.81 & \textbf{89.33} & 45.34 & 17.80 & 73.46 & 63.67 & \textbf{34.25} & 58.52 \\
\textbf{MoSLoRA-E} & Euclidean & 87.15 & 87.19 & 42.23 & 16.20 & 70.77 & 61.67 & 31.89 & 56.73 \\
\rowcolor{lightgray!29} \textbf{MoSLoRA (Ours)} & Mixture & \textbf{87.63} & 88.23 & \textbf{47.76} & \textbf{18.00} & \textbf{74.62} & \textbf{64.33} & 31.50 & \textbf{58.87} \\
\bottomrule
\end{tabular}
\end{small}
\end{table*}
\subsection{Tuning Dynamics}
\textbf{Curvature Dynamics of Distinct Space Experts.}
In our \textbf{MoSLoRA}, the curvature parameter $\kappa$, different from other capacity parameters $\theta$, serving as a key geometric property that simultaneously characterizes both the input space and the model's embedding space. As shown in Figure~\ref{fig:curvature_dynamic_layer}, we track the evolution of embedding spaces for geometric experts across model layers. Distinct colors denote different space types, with varying intensities representing spaces with different curvatures.
To provide stable Euclidean embeddings, two Euclidean experts with fixed curvature were included in all experiments. During training, the geometric experts progressively selected the curvature spaces that minimized the loss, and after approximately 6k–8k steps these selections stabilized, suggesting that the model had automatically identified the optimal mixture of embedding spaces under the given training configuration. We also observed interesting dynamic behaviors, such as recurrent switching across curvature spaces and transitions between geometric spaces. For instance, in Fig.~\ref{fig:image4}, one spherical expert in the 20th layer temporarily shifted into the hyperbolic space around step 4k, but subsequently returned to the spherical space and ultimately converged to a positive curvature near 0.2. These dynamics indicate that the model can adaptively adjust combinations of geometric experts across spaces, highlighting the flexibility and stability of our \textbf{MoS} architecture as a foundation for \textbf{MoSLoRA}. Furthermore, layer-wise trends reveals that lower layers, particularly the first layer, exhibit more stable but slower convergence compared to higher layers. This aligns with observations in MoE-related literature \citep{dai-etal-2024-deepseekmoe,muennighoff2024olmoe}: early layers primarily capture task-agnostic token-level information, leading to smoother geometry selection, while higher layers encode task-specific information, resulting in more pronounced cross-space transitions and dynamic curvature patterns, thereby underscoring the necessity of incorporating non-Euclidean geometry for downstream tasks.

\textbf{Optimizer for Space Experts.} We observed that when using a unified optimizer, the curvature $\kappa$ fail to adjust alongside other parameters, resulting in nearly static values and relatively poor task performance (in Table~\ref{table:optim_split}). Therefore, we assign independent optimizers and learning rates to $\kappa$ of all geometric experts, enabling the model to adjust latent geometric spaces more flexibly without being constrained by the optimization trajectory of other capacity-related parameters. Empirically, this setting leads to improved performance on downstream tasks.

\textbf{Token Frequency Distribution.} To further quantify and analyze the underlying geometric relationships from the perspective of token-level statistics, in Figure~\ref{fig:token_freq_norm}, we examine the token norm and frequency distributions produced by the tokenizers of Qwen2.5 models. We observe that the token frequency distribution exhibits a scale-free behavior, reflecting the latent hyperbolic characteristics of the underlying training corpus. Meanwhile, the token-norm distribution shows an exponential growth pattern on the right tail, whereas the left side deviates from such behavior. Prior work has partially observed similar phenomena \cite{yang2024hyperbolic}
, but their analysis focuses primarily on hyperbolic space, which is particularly well-suited for modeling power-law structures—yet less capable of capturing the broader range of non-Euclidean geometries that emerge across different contexts and domains.

\textbf{Geometric Mapping Efficiency.} In Figure~\ref{fig:efficiency_mapping}, we compare the runtime on GPUs between our proposed lightweight space mapping method and the conventional approach based on exp-log mappings. Figure~\ref{fig:image1} further illustrates how computation time scales with the increasing FFN dimension, while Figure~\ref{fig:image2} shows the corresponding trend related to the number of model layers. The results show that our method achieves significant speedups up to 4$\times$ over the standard exp–log scheme, and this relative acceleration remains consistent as the network depth increases and the embedding dimension grows. Refer to Appendix~\ref{appdix:efficiency} for detailed memory overhead and time comparison.

\subsection{Ablation Study}
\textbf{Different Space-Mix of Experts.} To obtain a more fine-grained understanding of how different geometric expert mixture configuration affect performance, we conducted an ablation study by restricting \textbf{MoSLoRA} to use only a single type of expert. For example, in the hyperbolic-only setting, each expert is initialized with a negative curvature (e.g., –1), and during the training stage, curvatures are learnable so that the model can dynamically adjust the underlying embedding space for inputs. All other configurations follow our default setup, including the unified framework, lightweight routing strategy, top-4 out of 8 expert selection, and the auxiliary load-balancing loss. As shown in Table~\ref{tb:different_mixture}, on average, \textbf{MoSLoRA} with mixed-geometric experts consistently outperforms all single-space variants. Nevertheless, specific single-space configurations achieved strong results on certain datasets. For instance, the hyperbolic-only variant \textbf{MoSLoRA-H} achieves accuracies of 17.80\% on MATH500, 34.25\% on AQuA, and 89.33\% on MRPC, matching or surpassing all competing methods, suggesting that hyperbolic embeddings are particularly well-suited for these benchmarks. The strong performance of hyperbolic experts on MATH500 further indicates their ability to generalize effectively in zero-shot settings to previously unseen reasoning problems. On the other hand, the Euclidean variant \textbf{MoSLoRA-E} outperforms others on CoLA (87.15\%), while the spherical variant \textbf{MoSLoRA-S} achieves the best results on SVAMP (64.33\%), highlighting that certain datasets benefit more from specific geometries. Together, these findings confirm the necessity of combining multiple spaces, as \textbf{MoSLoRA} effectively integrates the complementary strengths of different embedding geometries to achieve superior overall performance.


\section{Conclusion}
This paper first introduces MoS, a unified framework that integrates distinct geometric spaces and enables flexible transformations among three constant-curvature spaces: Hyperbolic, Euclidean, and Spherical spaces. Building on this framework, we further propose MoSLoRA, an efficient fine-tuning technique for LLMs that combines mixture-of-space experts. This design allows LLMs to dynamically adjust the curvature of each expert’s underlying space during fine-tuning and to flexibly reconfigure combinations of different spaces based on the input. In addition, to address the computational overhead of exponential and logarithmic operations commonly adopted in existing non-Euclidean models, we develop a lightweight routing and space-mapping strategy that improves the efficiency of space transitions. Experimental results demonstrate that our approach consistently outperforms strong baselines and provides new insights into geometric representation learning. Nevertheless, further investigation can assess its applicability to industrial-scale settings, as well as to reinforcement learning frameworks.

\section*{Impact Statement}
This paper presents work whose goal is to advance the field of Machine Learning. There are many potential societal consequences of our work, none which we feel must be specifically highlighted here.




\bibliography{example_paper}
\bibliographystyle{icml2026}

\newpage
\appendix
\onecolumn

\section{Training Details}

\subsection{Training hyperparameters}
\label{training_hyperparams}
Table \ref{table:hyperparameters_for_MoSLoRA}  presents the hyperparameters used to fine-tune the models with MoSLoRA on language understanding, commonsense reasoning and mathematical reasoning. The same hyperparameter settings are applied across all tasks. Each experiment is conducted independently, with a single run for each model. The final model trained is used for evaluation. For the baseline methods, the same hyperparameter configuration is reused. 

\begin{table}[!ht]
\centering
\caption{Hyperparameters for MoSLoRA.}
\begin{small}
\begin{tabular}{ll}
\toprule
\textbf{Hyperparameter} & \textbf{Value}\\
\midrule
Num Train Epoch & 3 \\
Optimizer & AdamW  \\
Weight Decay & 0.01 \\
Warmup Ratio & 0.1 \\
Learning Rate & $3 \times 10^{-4}$  \\
Projection Scaling $\gamma$ & 0.001 \\
Target Modules & \texttt{gate\_proj}, \texttt{down\_proj}, \texttt{up\_proj}\\
\bottomrule
\end{tabular}
\end{small}
\label{table:hyperparameters_for_MoSLoRA}
\end{table}

\subsection{Statistics of the language understanding dataset}
We conduct experiments using a subset of the General Language Understanding Evaluation (GLUE) dataset \citep{wang2018glue}, a benchmark for for training, evaluating, and analyzing natural language understanding systems. Specifically, we selected datasets Microsoft Research Paraphrase Corpus (MRPC) \citep{dolan2005automatically}, Corpus of Linguistic Acceptability (CoLA) \citep{warstadt-etal-2019-neural} and Recognizing Textual Entailment (RTE)~\citep{bowman-etal-2015-large}.
As shown in Table \ref{table:language_understanding_statistics}, this dataset consists of consist of various training and testing examples, each designed to evaluate specific linguistic tasks,including semantic equivalence and grammatical acceptability.

\begin{table}[!ht]
\centering
\caption{The detailed statistics of language understanding datasets.}
\begin{small}
\begin{tabular}{llll}
\toprule
\textbf{Dataset} & \textbf{Train} & \textbf{Test} & \textbf{Task Description} \\
\midrule
MRPC & 3,668 & 1,725 & Determine if a pair of sentences are semantically equivalent  \\
CoLA & 8,551 & 1,043 & Evaluate the grammatical acceptability of English sentences \\
RTE & 2,490 & 277 & Recognize if a hypothesis is entailed by text \\
\bottomrule
\end{tabular}
\end{small}
\label{table:language_understanding_statistics}
\end{table}

\subsection{Statistics of mathematical reasoning datasets}
As illustrated in the Table \ref{table:mathematical_reasoning_statistics}, we have constructed a mathematical reasoning training set consisting of a mix of four datasets, totaling 13,262 examples. These datasets include GSM8K \citep{cobbe2021training}, MAWPS \citep{koncel2016mawps}, SVAMP \citep{patel2021nlp}, and subset of AQuA \citep{DBLP:journals/corr/LingYDB17}, each focusing on different aspects of mathematical reasoning. To control the contribution of AQuA, we do not use its full training split; instead, we uniformly sample 4,000 examples from AQuA’s training set using a fixed random seed (42). 
Additionally, we have incorporated MATH500 \citep{lightman2023let} into the test set to further evaluate the model's performance.
\begin{table}[!ht]
\centering
\caption{The detailed statistics of mathematical reasoning datasets.}
\begin{small}
\begin{tabular}{lll}
\toprule
\textbf{Dataset} & \textbf{Data Number} & \textbf{Task Type} \\
\midrule
Train & 13,262 & Mixed \\
\midrule
Test &  &  \\
GSM8K & 1,319 & Question-Answering \\
MAWPS & 520 & Question-Answering \\
SVAMP & 300 & Question-Answering \\
MATH500 & 500 & Question-Answering \\
AQuA & 254 & Option \\
\bottomrule
\end{tabular}
\end{small}
\label{table:mathematical_reasoning_statistics}
\end{table}

\subsection{Statistics of commonsense reasoning datasets}
Also, to extend the scope of our experiments beyond mathematical reasoning and language understanding, we further evaluate our approach on commonsense reasoning tasks. As shown in Table~\ref{table:commonsense_reasoning_statistics}, we construct the commonsense reasoning training set using two widely adopted multiple-choice benchmarks: OpenBookQA (OBQA) ~\citep{mihaylov-etal-2018-suit} and CommonsenseQA (CSQA)~\citep{talmor-etal-2019-commonsenseqa}. For training, we use the official training splits of both datasets. Evaluation is conducted on their respective validation and test sets to ensure consistency with prior work.

\begin{table}[!ht]
\centering
\caption{The detailed statistics of language understanding datasets.}
\begin{small}
\begin{tabular}{llll}
\toprule
\textbf{Dataset} & \textbf{Train} & \textbf{Test}& \textbf{Task Description} \\
\midrule
OBQA & 4,957 & 500   & Answer science questions using provided facts plus commonsense\\
CSQA & 9,742 & 1,221 & Choose correct option requiring diverse commonsense knowledge\\
\bottomrule
\end{tabular}
\end{small}
\label{table:commonsense_reasoning_statistics}
\end{table}

\section{Gradient bound Analysis}
\label{appendix:grad_bound}
The following presents a gradient analysis of our MoSLoRA framework, demonstrating that the gradients in our design remain bounded.
\paragraph{Bound on the gradient w.r.t.\ $u$.}
Let \(u=\phi(z)\) and define the lifting coordinate
\[
a_{\kappa}(u)=\sqrt{\operatorname{sgn}(-\kappa)\,\|u\|^{2}+\varphi(\kappa)},
\qquad
\varphi(\kappa)=
\begin{cases}
1/|\kappa|, & \kappa\neq 0,\\[2pt]
0, & \kappa=0~.
\end{cases}
\]
A direct differentiation gives
\begin{equation}
\nabla_{u} a_{\kappa}(u)
=\frac{\operatorname{sgn}(-\kappa)\,u}
{\sqrt{\operatorname{sgn}(-\kappa)\,\|u\|^{2}+\varphi(\kappa)}},
\qquad
\bigl\|\nabla_{u} a_{\kappa}(u)\bigr\|
=\frac{\|u\|}{\sqrt{\operatorname{sgn}(-\kappa)\,\|u\|^{2}+\varphi(\kappa)}}.
\label{eq:grad-u}
\end{equation}

\noindent\textbf{Case \(\kappa<0\) (hyperbolic).}
Here \(\operatorname{sgn}(-\kappa)=1\) and \(\varphi(\kappa)=1/|\kappa|\). Hence
\[
\bigl\|\nabla_{u} a_{\kappa}(u)\bigr\|
=\frac{\|u\|}{\sqrt{\|u\|^{2}+1/|\kappa|}}
\le 1 \quad \text{for all }u,
\]
so \(a_{\kappa}\) is globally \(1\)-Lipschitz in \(u\).

\noindent\textbf{Case \(\kappa>0\) (spherical).}
Now \(\operatorname{sgn}(-\kappa)=-1\) and the domain requires \(\|u\|^{2}\le 1/|\kappa|\).
Let \(R=\sup\|u\|<1/\sqrt{|\kappa|}\), which can be enforced by a bounded activation
together with scaling/projection. Then, from \eqref{eq:grad-u},
\[
\bigl\|\nabla_{u} a_{\kappa}(u)\bigr\|
=\frac{\|u\|}{\sqrt{1/|\kappa|-\|u\|^{2}}}
\le \frac{R}{\sqrt{1/|\kappa|-R^{2}}}
=: C_{u}(\kappa,R) < \infty .
\]

Therefore, as long as \(u\) stays at a fixed margin from the boundary
(e.g., \(R=(1-\varepsilon)/\sqrt{|\kappa|}\) with \(\varepsilon>0\)),
the gradient is uniformly bounded.

\section{Scale Equivariance of the Unified Geometric Expert}
\label{appdix:equivar}

We consider the full non-Euclidean feed-forward mapping defined by the composition
\[
F_{\kappa}(x) := \rho_{\kappa}\!\left(G\!\left(W,\,x\right)\right),
\]
where the mapping consists of the inverse stereographic projection, the unified geometric expert transformation, and the
final projection back to Euclidean space.
We show that this mapping is \emph{scale-equivariant} under a coupled
rescaling of the input and curvature.

\paragraph{Claim.}
For any scalar \(\gamma>0\) and nonzero curvature \(\kappa \neq 0\),
\[
F_{\kappa}(x) \;=\; \frac{1}{\gamma}\,F_{\kappa / \gamma^{2}}(\gamma x).
\]

\paragraph{Proof.}
We denote the rescaled curvature by
\[
\tilde{\kappa} := \frac{\kappa}{\gamma^{2}}.
\]

\medskip
\noindent\textbf{Step 1: Scaling behavior of the space-like component (Eq.~(5)).}
The space-like component produced by the inverse stereographic projection is
\[
s_{\kappa}(x) = \frac{2x}{1+\kappa\|x\|^{2}}.
\]
Evaluating this expression at the rescaled input \(\gamma x\) with curvature \(\tilde{\kappa}\),
we obtain
\[
\begin{aligned}
s_{\tilde{\kappa}}(\gamma x)
&= \frac{2(\gamma x)}{1+\tilde{\kappa}\|\gamma x\|^{2}} \\
&= \frac{2\gamma x}{1+(\kappa/\gamma^{2})\,\gamma^{2}\|x\|^{2}} \\
&= \frac{2\gamma x}{1+\kappa\|x\|^{2}} \\
&= \gamma\, s_{\kappa}(x).
\end{aligned}
\]
Applying the linear transformation \(W\) yields
\[
s'_{\tilde{\kappa}}(\gamma x) = W s_{\tilde{\kappa}}(\gamma x)
= \gamma\, W s_{\kappa}(x)
= \gamma\, s'_{\kappa}(x).
\]

\medskip
\noindent\textbf{Step 2: Scaling behavior of the time-like component.}
The time-like coordinate is defined as
\[
\xi' = \sqrt{\|s'\|^{2}\,\mathrm{sgn}(-\kappa) + \varphi(\kappa)},
\qquad
\varphi(\kappa)=\frac{1}{|\kappa|}.
\]
Note that
\[
\mathrm{sgn}(-\tilde{\kappa})=\mathrm{sgn}(-\kappa),
\qquad
\varphi(\tilde{\kappa})=\frac{1}{|\tilde{\kappa}|}
= \frac{\gamma^{2}}{|\kappa|} = \gamma^{2}\varphi(\kappa).
\]
Therefore,
\[
\begin{aligned}
\xi'_{\tilde{\kappa}}(\gamma x)
&= \sqrt{\|s'_{\tilde{\kappa}}(\gamma x)\|^{2}\,\mathrm{sgn}(-\kappa)
+ \varphi(\tilde{\kappa})} \\
&= \sqrt{\gamma^{2}\|s'_{\kappa}(x)\|^{2}\,\mathrm{sgn}(-\kappa)
+ \gamma^{2}\varphi(\kappa)} \\
&= \sqrt{\gamma^{2}\bigl(\|s'_{\kappa}(x)\|^{2}\,\mathrm{sgn}(-\kappa)
+ \varphi(\kappa)\bigr)} \\
&= \gamma\, \xi'_{\kappa}(x).
\end{aligned}
\]

\medskip
\noindent\textbf{Step 3: Scaling behavior of the projection.}
The final projection is given by
\[
\rho_{\kappa}(\xi', s') =
\frac{s'}{1+\sqrt{|\kappa|}\,\xi'}.
\]
Substituting the scaled quantities yields
\[
\begin{aligned}
F_{\tilde{\kappa}}(\gamma x)
&= \rho_{\tilde{\kappa}}\!\left(\xi'_{\tilde{\kappa}}(\gamma x), s'_{\tilde{\kappa}}(\gamma x)\right) \\
&= \frac{\gamma\, s'_{\kappa}(x)}
{1+\sqrt{|\tilde{\kappa}|}\,\gamma\,\xi'_{\kappa}(x)} \\
&= \frac{\gamma\, s'_{\kappa}(x)}
{1+\sqrt{|\kappa|}\,\xi'_{\kappa}(x)} \\
&= \gamma\, \rho_{\kappa}\!\left(\xi'_{\kappa}(x), s'_{\kappa}(x)\right) \\
&= \gamma\, F_{\kappa}(x).
\end{aligned}
\]
Dividing both sides by \(\gamma\) yields
\[
F_{\kappa}(x) = \frac{1}{\gamma}\,F_{\kappa / \gamma^{2}}(\gamma x),
\]
which completes the proof.
\hfill\(\square\)

\paragraph{Remark.}
The above scale-equivariance property does \emph{not} hold if the curvature
\(\kappa\) is kept fixed while scaling the input \(x\).
This is due to the non-homogeneous dependence on \(\kappa\) in both
the inverse stereographic projection denominator and the curvature-dependent
offset \(\varphi(\kappa)\).

\section{Main Results on Gemma2}
\label{main_result_gemma2}
To verify that the proposed method is not limited to a specific
architectural design, we evaluate its performance across multiple
parameter-efficient fine-tuning variants under the same backbone model Gemma2-2b-it, as is shown in Table~\ref{tab:main_results_extended_gemma2-2b}.
This allows us to examine the robustness and generality of our approach
across different architectural configurations.

\begin{table}[t]
\centering
\caption{Comparison of parameter-efficient fine-tuning methods across reasoning and language understanding benchmarks.
All results are reported on the same backbone model.}
\small
\begin{tabular}{l|cccccccc|c}
\toprule
\textbf{Method}  & \textbf{GSM8k} & \textbf{MATH500} & \textbf{MAWPS} & \textbf{SVAMP} & \textbf{AQuA} & \textbf{MRPC} & \textbf{CoLA} & \textbf{RTE} & \textbf{Avg.} \\
\midrule
LoRA       & \textbf{52.24} & 13.80 & 71.33 & 65.19 & \textbf{32.68} & 89.39 & 84.66 & 90.25 & 62.44 \\
AdaLoRA    & 46.70 & 7.20  & 64.23 & 62.67 & 20.08 & 84.17 & 83.03 & 87.36 & 56.93 \\
DoRA       & 48.60 & 11.20 & 71.67 & 64.62 & 31.89 & \textbf{89.45} & 85.33 & 89.89 & 61.58 \\
MELoRA     & 48.90 & 11.80 & 63.85 & 70.00 & 28.35 & 88.75 & 84.08 & 88.81 & 60.57 \\
HypLoRA    & 45.34 & 10.80 & 64.23 & 66.33 & 27.56 & 87.25 & 84.08 & 87.36 & 59.12 \\
HydraLoRA  & \textbf{51.18} & \textbf{14.80} & 65.38 & 71.33 & 28.35 & 88.58 & \textbf{86.29} & \textbf{90.97} & 62.11 \\
\midrule
\textbf{MoSLoRA} & \textbf{47.53} & \underline{\textbf{14.40}} & \textbf{81.35} & \textbf{73.00} & \underline{\textbf{31.50}} & \underline{\textbf{88.63}} & \underline{\textbf{85.43}} & \underline{\textbf{89.17}} & \textbf{63.88} \\
\bottomrule
\end{tabular}
\label{tab:main_results_extended_gemma2-2b}
\end{table}

\section{Efficiency}
\label{appdix:efficiency}

We evaluate the computational efficiency of the proposed MoS framework
by comparing it against the standard exponential/logarithmic (Exp/Log)
mapping.
For each method, we report the total execution time, as well as the
average runtime and memory consumption for both forward and backward
passes.
All measurements are conducted under identical hardware and batch-size
settings to ensure fair comparison.

As shown in Table~\ref{tab:efficiency_mapping}, MoS substantially reduces
the overall conversion time compared to the Exp/Log mapping.
In particular, MoS achieves a total runtime of $57.46$ ms, less than
half of the $135.08$ ms required by Exp/Log.

In terms of memory usage, MoS demonstrates significantly improved
efficiency during the forward pass, consuming approximately $277$ MB,
compared to $566$ MB for Exp/Log.
During the backward pass, MoS requires around $324$ MB of memory, which
is comparable to the baseline.
Overall, these results indicate that MoS not only accelerates geometric
transformations but also maintains competitive or lower memory
consumption, especially in the forward pass, thereby offering superior
computational efficiency.

\begin{table}[t]
\centering
\caption{Runtime and memory consumption comparison between MoS and
Exp/Log geometric mappings.
Results are reported for both forward and backward passes.}
\small
\setlength{\tabcolsep}{6pt}
\begin{tabular}{lccccc}
\toprule
\textbf{Method} & \textbf{Total Time (ms)} & \textbf{Phase} &
\textbf{Avg Time (ms)} & \textbf{Avg Memory (MB)} \\
\midrule
\multirow{2}{*}{MoS}
 & \multirow{2}{*}{\textbf{57.4}}
 & Forward & 2.1 & 277.2 \\
 &  & Backward & 2.1 & 324.0 \\
\midrule
\multirow{2}{*}{Exp/Log}
 & \multirow{2}{*}{135.0}
 & Forward & 2.6 & 565.6 \\
 &  & Backward & 8.6 & 396.0 \\
\bottomrule
\end{tabular}
\label{tab:efficiency_mapping}
\end{table}


\end{document}